\documentclass[10pt,twocolumn,letterpaper]{article}

\usepackage[pagenumbers]{iccv} 

\definecolor{iccvblue}{rgb}{0.21,0.49,0.74}
\usepackage[pagebackref,breaklinks,colorlinks,allcolors=iccvblue]{hyperref}

\usepackage{graphicx}
\usepackage{amsmath}
\usepackage{amssymb}
\usepackage{booktabs}
\usepackage{bm}
\usepackage{multirow}
\usepackage{float}

\usepackage[normalem]{ulem}
\usepackage{color, colortbl}
\definecolor{cGreen}{RGB}{78,134,86}

\def\paperID{1635} 
\def\confName{ICCV}
\def\confYear{2025}

\title{LTM3D: Bridging Token Spaces for Conditional 3D Generation with Auto-Regressive Diffusion Framework}

\author{
{Xin Kang{$^{1}$}}\footnotemark[1] ~ Zihan Zheng{$^{1}$}\footnotemark[1] ~ Lei Chu{$^{2}$} ~ Yue Gao{$^{2}$} ~ Jiahao Li{$^{2}$} ~ Hao Pan{$^{3}$} ~ Xuejin Chen{$^{1}$} ~ Yan Lu{$^{2}$}\\
\normalsize
$^{1}$\, University of Science and Technology of China ~  $^{2}$\, Microsoft Research Asia ~$^{3}$\, Tsinghua University\\
}

\begin{document}

\twocolumn[{%
\renewcommand\twocolumn[1][]{#1}%
\maketitle
\vspace{-8mm}
\centering
\includegraphics[width=1.0\linewidth]{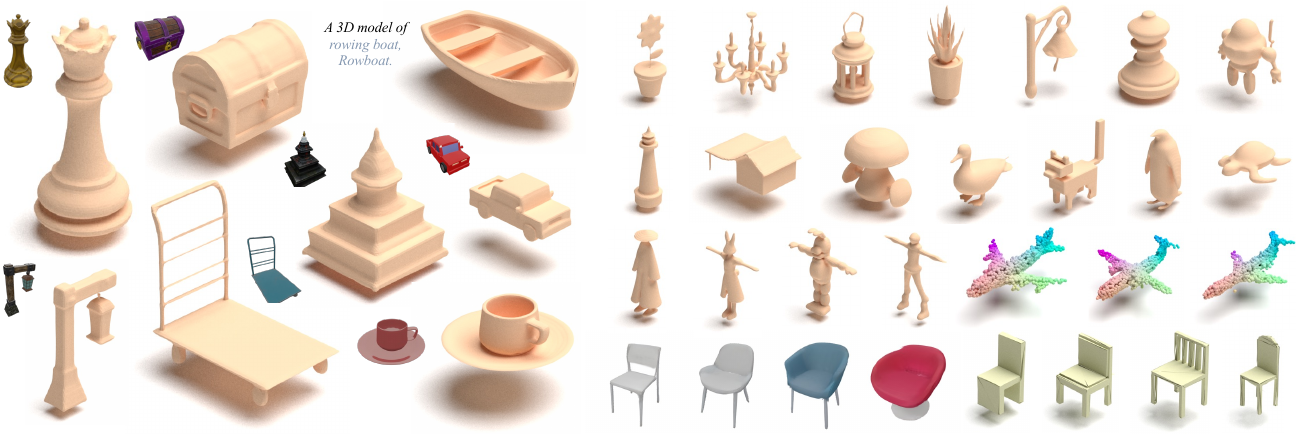}
\captionof{figure}{LTM3D is a conditional 3D shape generation framework that supports image and text inputs, as well as multiple 3D data representations, including signed distance fields, 3D Gaussian Splatting, point clouds, and meshes.}
\vspace{4mm}
\label{fig:teaser}
}]

\renewcommand{\thefootnote}{*}
\footnotetext[1]{This open-source research project began in March 2024 and was conducted during Xin and Zihan’s internship at Microsoft Research Asia.}

\begin{abstract}
We present LTM3D, a Latent Token space Modeling framework for conditional 3D shape generation that integrates the strengths of diffusion and auto-regressive (AR) models. While diffusion-based methods effectively model continuous latent spaces and AR models excel at capturing inter-token dependencies, combining these paradigms for 3D shape generation remains a challenge. To address this, LTM3D features a Conditional Distribution Modeling backbone, leveraging a masked autoencoder and a diffusion model to enhance token dependency learning. Additionally, we introduce Prefix Learning, which aligns condition tokens with shape latent tokens during generation, improving flexibility across modalities. We further propose a Latent Token Reconstruction module with Reconstruction-Guided Sampling to reduce uncertainty and enhance structural fidelity in generated shapes. Our approach operates in token space, enabling support for multiple 3D representations, including signed distance fields, point clouds, meshes, and 3D Gaussian Splatting. Extensive experiments on image- and text-conditioned shape generation tasks demonstrate that LTM3D outperforms existing methods in prompt fidelity and structural accuracy while offering a generalizable framework for multi-modal, multi-representation 3D generation.

\end{abstract}
\section{Introduction} \label{section:introduction}

In the era of large models~\cite{achiam2023gpt, touvron2023llama, kirillov2023segment, oquab2023dinov2, alayrac2022flamingo, rombach2022high}, the concept of token space has emerged as pivotal for managing cross-modality interactions, significantly advancing the development of Artificial General Intelligence. This importance is underscored by recent breakthroughs in various fields, such as CLIP~\cite{radford2021learning}, DALL-E~\cite{ramesh2021zero} and CLAY~\cite{zhang2024clay}, which highlight the role of token space in handling diverse modalities. In 3D generation, tokenization converts complex shape geometries into structured, manageable tokens that machine learning models can efficiently process, enabling the capture of intricate relationships within 3D data.

Existing methods~\cite{cheng2023sdfusion, zhang2024gaussiancube, zhang20233dshape2vecset, jun2023shap, zhang2024clay, siddiqui2024meshgpt, chen2024meshxl, nash2020polygen} for conditional shape generation can be broadly categorized into two primary types: diffusion-based models and auto-regressive (AR) models. 
Diffusion-based models, leveraging UNet-based architectures~\cite{cheng2023sdfusion, zhang2024gaussiancube} or Transformer-based architectures~\cite{zhang20233dshape2vecset, jun2023shap, zhang2024clay}, effectively model data distributions by progressively reversing a Markovian noising process. This step-by-step denoising approach enables them to capture complex data distributions more effectively than traditional generative models~\cite{rombach2022high}.
However, prior work~\cite{ma2019flowseq} suggests that as token sequence length increases, modeling the joint distribution over an exponentially large latent space becomes increasingly challenging. This hinders the model’s ability to capture intricate shape structures in high-dimensional 3D data.
In the meanwhile, AR methods excel at capturing inter-token dependencies by decomposing complex joint distributions into products of conditional probabilities. However, their performance is limited by the constraints of a quantized latent space, which can result in the loss of fine-grained geometric details.
Existing shape encoding methods prioritize preserving intricate structural details, resulting in a high-dimensional, continuous latent space. While diffusion models are well-suited for modeling such continuous spaces and AR models effectively capture high-dimensional joint distributions, integrating the strengths of both approaches for 3D latent space modeling remains an open challenge.


Inspired by the success of MAR~\cite{li2024autoregressive} in image generation, we introduce \textit{LTM3D}, a \textit{Latent Token space Modeling framework} that leverages the strengths of both diffusion and auto-regressive models for conditional 3D shape generation.
At its core, \textit{LTM3D} features a \textit{conditional distribution modeling backbone} that integrates a masked autoencoder to capture token dependencies between condition and shape tokens, along with a diffusion model to learn the conditional token distribution. However, the domain gap between multi-modal condition tokens and the shape tokens, as well as the uncertainty in generated tokens, still hinders the performance of conditional 3D shape generation.
To address these challenges, we propose two modules: \textit{prefix learning} and \textit{latent token reconstruction} module. Unlike previous works~\cite{zhao2024michelangelo} that align the shape token space with the condition token space during the shape encoding process, our \textit{prefix learning } module performs the alignment during the generation process. This decouples the encoding process from generation, allowing the adoption of state-of-the-art encoding methods and making it more flexible to accommodate other condition types, such as audio data.
Moreover,  \textit{latent token reconstruction} module tackles the problem of token uncertainty by mapping condition tokens to shape latent tokens, serving as priors to guide the sampling process. Through our \textit{reconstruction-guided sampling} strategy, these reconstructed latent tokens provide additional shape guidance, reducing uncertainty and improving alignment between generated results and input conditions.
Operating in token space, \textit{LTM3D} serves as a versatile framework that can support multi-modal conditional prompts and integrate various shape encoders for diverse shape representations, enabling faithful and high-fidelity conditional generation. As shown in Fig.~\ref{fig:teaser}, \textit{LTM3D} seamlessly supports 3D shapes across various representations, eliminating the need for separate generation backbones for different types of 3D data.

The key contributions can be summarized as follows.
\begin{itemize}
    \item We propose LTM3D, a general latent token space modeling framework for conditional shape generation capable of supporting diverse 3D data representations.  
    \item We propose a \textit{Prefix Learning} module to bridge the domain gap between condition tokens and shape latent tokens for faithful shape generation.
    \item We design a \textit{Latent Token Reconstruction} module and a \textit{Reconstruction Guided Sampling} strategy to improve the structural fidelity of generated shapes.
    \item  We achieve state-of-the-art results on both image- and text-conditioned 3D shape generation tasks.
\end{itemize}

\begin{figure*}[!t]
  \centering
  \includegraphics[width=1.0\linewidth]{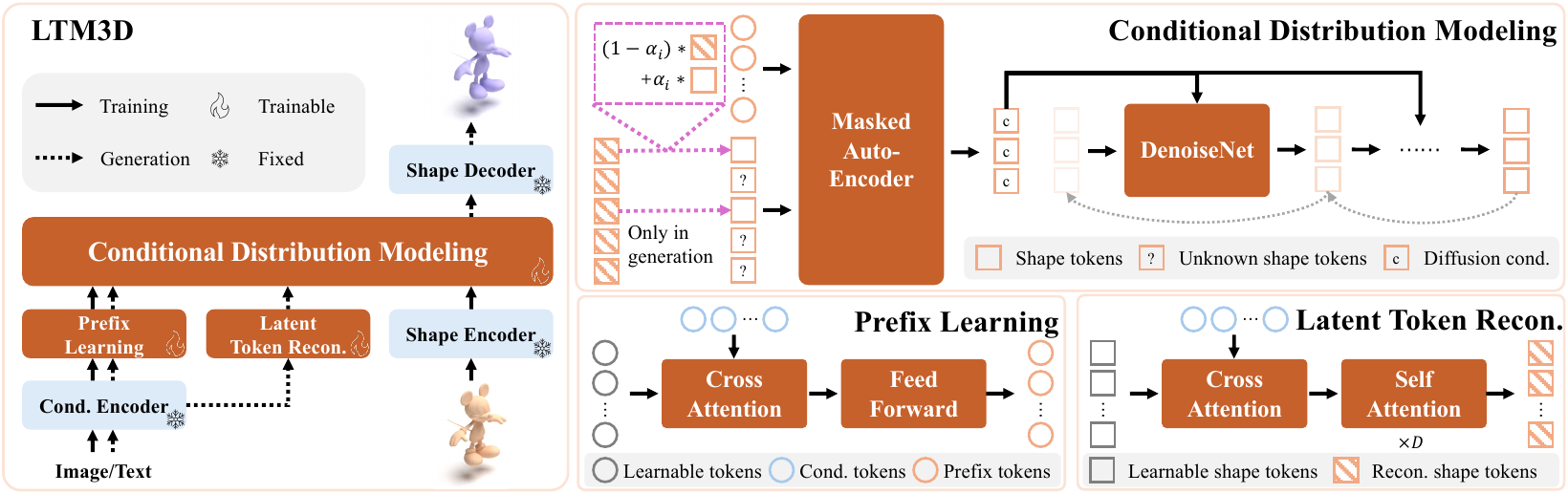}
   \vspace{-5mm}
  \caption{Our framework LTM3D consists of three components. In Conditional Distribution Modeling, we implement an auto-regressive diffusion backbone that learn inter-token dependencies with a Masked Auto-Encoder and learn conditional token distributions with an MLP-based DenoiseNet.
  In Prefix Learning, we use a cross-attention layer and a feed-forward layer to project condition tokens into prefix tokens. In Latent Token Reconstruction, we use a cross-attention layer and $D$ self-attention layers to reconstruct shape tokens from input condition tokens. Note that Latent Token Reconstruction module is trained separately with a reconstruction loss.}
  \label{fig:pipeline}
  \vspace{-5mm}
\end{figure*}


\section{Related Work} 

\noindent
\textbf{Image-driven approaches.}
Recent advances in 2D generative models have enabled the creation of high-quality images with impressive details, providing a foundation for the image-driven methods, which use 2D generation priors to guide 3D generation. DreamFusion \cite{poole2022dreamfusion} pioneered this with Score Distillation Sampling (SDS), leveraging powerful 2D models like Imagen \cite{saharia2022photorealistic} and Stable Diffusion \cite{zhang2023adding} for 3D generation through NeRF \cite{mildenhall2021nerf} optimization. However, DreamFusion struggles with multi-view consistency, resulting in reduced 3D detail quality. Following methods \cite{lin2023magic3d,tang2023make,huang2023dreamtime,yu2023points,zhu2023hifa,zhang2023adding,wu2024hd,chen2024text} extend SDS to other 3D formats, such as point clouds and 3D Gaussians \cite{kerbl20233d}, often using two-stage training to improve quality. Meanwhile, Zero-1-to-3 \cite{ramesh2021zero} embeds multi-view information directly in the image generation module, using a fine-tuned image diffusion model to create consistent multi-view outputs that are used to optimize a coherent 3D model.
Despite these successes in improving the generation faithfulness, inconsistencies across multi-view images still exist and often lead to blurriness and artifacts. In this paper, we focus on 3D-driven approaches that utilize 3D data during training, thus enabling high quality shape generation.

\noindent
\textbf{3D-driven approaches.}
3D-driven approaches leverage 3D datasets for training to capture fine geometric details, offering higher shape quality compared to image-driven methods. Current 3D-driven methods fall into two main categories: diffusion-based and auto-regressive (AR) models.
Diffusion-based models are either UNet-based or transformer-based. UNet-based diffusion is suited for volumetric data, like GaussianCube \cite{zhang2024gaussiancube}, which maps 3D Gaussians onto a volumetric grid, and SDFusion \cite{cheng2023sdfusion}, which compresses signed distance fields into latent volumes. Transformer-based diffusion models are suitable for latent sequences, such as point cloud embeddings \cite{nichol2022point} and learnable latent vector sets \cite{zhang20233dshape2vecset,zhang2024clay}.
AR models excel at handling complex sequential data like meshes. MeshGPT \cite{siddiqui2024meshgpt} and MeshXL \cite{chen2024meshxl} tokenize mesh vertices and faces into sequences and adopt a transformer backbone \cite{vaswani2017attention} to capture token-wise dependencies.
In contrast to these methods, our LTM3D is a general conditional shape generation framework that supports diverse 3D latent representations.

\noindent
\textbf{Faithfulness in 3D generation.}
A key challenge in 3D generation is to ensure consistency between the input prompts (image or text) and the generated 3D shapes. Recent work, Michelangelo \cite{zhao2024michelangelo}, enhances faithfulness by aligning 3D latent space with the CLIP latent space before diffusion training. In this work, we introduce a novel \textit{Prefix Learning} module to align condition tokens with latent tokens for faithful conditioned generation, coupled with a \textit{Reconstruction Guided Sampling} strategy for more deterministic sampling, resulting in improved generation fidelity.


\section{Method} \label{section:method}

Given a 3D shape space $\mathcal{S}$ under a specific representation (e.g., signed distance function (SDF), point cloud, mesh, or 3D Gaussian Splatting (3DGS)) with its pretrained encoder $f_{enc}:\mathcal{S}\Rightarrow\mathcal{X}$ mapping the shape space into a latent token space $\mathcal{X}$ flattened from the encoded latent representation, our goal is to build a general conditional latent token generation framework that can handle multiple input conditions and subsequently decode them into 3D shapes.
As shown in Fig.~\ref{fig:pipeline}, our framework comprises three main components. In Sec.~\ref{method:method_PL}, we present prefix learning module, which aligns condition tokens with shape tokens during the generation process.
In Sec.~\ref{method:method_MAR}, we introduce our conditional distribution modeling backbone to effectively learn token dependencies and capture condition distributions.  In Sec.~\ref{method:recon_guide_sampling}, we describe latent token reconstruction module, which mitigates the impact of token uncertainty. 

\subsection{Prefix Learning}\label{method:method_PL}
Existing conditional generation methods~\cite{li2024autoregressive, zhang20233dshape2vecset, zhang2024clay} often directly use image or text tokens as conditioning inputs. However, the inherent domain gap between condition tokens (e.g., image or text representations) and the shape latent space can hinder the faithfulness of generated shapes to the input conditions, as highlighted in Michelangelo~\cite{zhao2024michelangelo}.

Unlike previous works ~\cite{zhao2024michelangelo}, which aim to address this domain gap during the encoding process, we propose to bridge the gap during the \textit{generation process}. By decoupling encoding from generation, our method is more flexible and adaptable to various input conditions while enabling the use of diverse powerful shape encoders.

Specifically, given condition tokens $\bm{C}$ encoded from input image or text description, we initialize a set of learnable tokens, $\bm{Q} = \{\bm{q}_1, \dots, \bm{q}_k\}$, as network parameters. 
These learnable tokens are processed through a cross-attention mechanism, where semantic information from the condition tokens $\bm{C}$ is integrated, followed by a feed-forward (FF) layer. The resulting prefix tokens $\bm{T}$ are computed as:
\begin{equation}
\bm{T} = {\rm FF}({\rm CrossAttn}(\bm{Q}, \bm{C}) + \bm{Q}).
\end{equation}

By aligning condition tokens to the shape latent space during generation, prefix learning ensures more seamless conditioning, which improves the consistency of generated shapes with their input prompts. Additionally, using prefix learning allows us to generalize conditions across diverse prompt types (image and text) while maintaining compatibility within the 3D shape generation framework. This results in a model capable of generating accurate shapes across diverse 3D representations, while also improving faithfulness to prompts and adaptability to various multi-modal input conditions.

In our implementation, we encode the input image via a pre-trained DinoV2~\cite{oquab2023dinov2} and encode the input text description using a pre-trained CLIP~\cite{radford2021learning} model.

\subsection{Conditional Distribution Modeling}\label{method:method_MAR}
Our latent token space modeling backbone is built upon the shape latent space and utilizes a masked auto-regressive (MAR)~\cite{li2024autoregressive} framework. It consists of two key components: a Masked Autoencoder (MAE) for learning inter-token dependencies and an MLP-based DenoiseNet for learning conditional token distributions.

\noindent \textbf{Inter-token Dependency Learning.} 
Previous methods~\cite{zhao2024michelangelo, zhang20233dshape2vecset, zhang2024gaussiancube,zhang2024clay} learn the conditional joint distribution by minimizing an objective formulated as $\text{D}(q(\bm{X}|\bm{T}), p(\bm{X}|\bm{T}))$, where $\bm{X}\in\mathcal{X}$ represents the shape tokens, $q(\bm{X}|\bm{T})$ is the estimated distribution, and $\text{D}(\cdot,\cdot)$ denotes a distance function between two distributions. However, as sequence length increases, directly optimizing this joint distribution becomes increasingly challenging due to the exponential complexity of inter-token dependencies~\cite{ma2019flowseq}. This can hinder the model's ability to capture intricate details in shape latent tokens. To address this issue, we factorize the joint conditional distribution into a product of conditional distributions as follows:
\[
p(\bm{X}|\bm{T}) = \prod_{i=1}^n p(\bm{x}_i | \bm{x}_{<i}, \bm{T}),
\]
where $p(\bm{x}_i | \bm{x}_{<i}, \bm{T})$ represents the conditional distribution of token $\bm{x}_i$ given the preceding tokens $\bm{x}_{<i}$ and the condition tokens $\bm{T}$. 
We follow MAE~\cite{he2022masked} by randomly shuffling the sequence order to help the model learn more diverse conditional dependencies.
This reformulation allows the optimization objective to be expressed as:
\begin{equation}
\text{min} \sum_{i=1}^n \text{D}(q(\bm{x}_i | \bm{x}_{<i}, \bm{T}),~p(\bm{x}_i | \bm{x}_{<i}, \bm{T})).
\end{equation}

To simplify optimization, we approximate $q(\bm{x}_i | \bm{x}_{<i}, \bm{T})$ as $q(\bm{x}_i | \bm{z}_i)$, where $\bm{z}_i=f(\bm{x}_{<i}, \bm{T})$ encodes the dependencies between  token $\bm{x}_i$, its preceding tokens $\bm{x}_{<i}$, and the condition tokens $\bm{T}$. 
As shown in Fig.~\ref{fig:pipeline}, we employ a Masked Auto-encoder (MAE) to learn inter-token dependencies as $\bm{z}_i=\text{MAE}(\bm{x}_{<i}, \bm{T})$.

\noindent\textbf{Next-token Distribution Modeling.} 
Our approach departs from traditional diffusion methods, which model $p(\bm{X}|\bm{T})$ as a joint distribution. Such methods can be difficult to optimize for complex, high-dimensional 3D data. Instead, we adopt the decomposition strategy used in MAR~\cite{li2024autoregressive}, where the factorized conditional distribution $p(\bm{x}_i|\bm{z}_i)$ is independently modeled through a DenoiseNet.
The DenoiseNet is optimized using the following objective function:
\begin{equation} \label{eq:diff_loss}
\mathcal{L}(\bm{x}_i, \bm{z}_i) = \mathbb{E}_{\epsilon, t}[\Vert \epsilon - \epsilon_\theta(\bm{x}_i|t,\bm{z}_i) \Vert^2], \end{equation}
where $\epsilon$ is a noise token sampled from a Gaussian distribution $\mathcal{N}(\bm{0}, \bm{I})$, $t$ represents the time step, and $\epsilon_\theta$ is the predicted noise token from the DenoiseNet parameterized by $\theta$. 
This factorized diffusion approach refines each token iteratively based on its inter-token dependencies, allowing for more accurate modeling of intricate 3D shapes.

\subsection{Latent Token Reconstruction} \label{method:recon_guide_sampling}
During the generation, we auto-regressively sample shape tokens for shape decoding.
In this process, the probabilistic distribution of the first several tokens, $p(\bm{x}_i|\bm{x}_{<i},\bm{T})$, often lacks determinism due to limited conditioning information, potentially leading to diverse sampled shapes that are not faithful to the input prompts. 
To ensure accurate generated shapes aligned with the input conditions, we design a \textit{latent token reconstruction} module that reconstructs shape tokens from the condition tokens, as illustrated Fig.~\ref{fig:pipeline}. 
This module comprises a cross-attention layer followed by $D$ self-attention blocks.
We first initialize a set of learnable tokens $\bm{Q}^S$ as the optimizable parameters, then generate reconstructed tokens $\bm{\hat{X}}$ following
\begin{equation}
\bm{\hat{X}} = {\rm SelfAttn}({\rm CrossAttn}(\bm{Q}^S, \bm{C}) + \bm{Q}^S).
\end{equation}

Given the reconstructed tokens $\bm{\hat{x}}_{<i}$ alongside the diffusion-sampled tokens $\bm{x}_{<i}^S$, we blend them through a linear combination weighted by $\alpha_i$, formulating the input sequence $\bm{x}_{<i}^F$ for the MAE network. The process for sampling the next token is formulated as:
\begin{equation} 
\begin{split}
\bm{x}_{<i}^F&=(1-\alpha_i)\cdot\bm{\hat{x}}_{<i}+\alpha_i\cdot\bm{x}_{<i}^S, \\
\bm{z}_i&={\rm MAE}(\bm{x}_{<i}^F, \bm{T}), \\
\bm{x}_i^S&={\rm Sample}(\bm{z}_i).
\end{split}
\end{equation}

By incorporating guidance from reconstructed tokens, the condition vector $\bm{z}_i$ can leverage more accurate blended tokens $\bm{x}_{<i}^F$, effectively reducing sampling uncertainty and enhancing shape alignment with input conditions.
\section{Experiment} \label{section:exp}

\begin{table*}
  \centering
  \begin{tabular}{l|cc|cc|cc|cc|c|c}
    \toprule
    \multirow{2}{*}{Methods} & \multicolumn{2}{c|}{ULIP\,($\uparrow$)} & \multicolumn{2}{c|}{CD\,($\downarrow$)} & \multicolumn{2}{c|}{EMD\,($\downarrow$)} & \multicolumn{2}{c|}{F-Score\,($\uparrow$)} & \multirow{2}{*}{P-FID\,($\downarrow$)} & \multirow{2}{*}{P-IS\,($\uparrow$)} \\ \cline{2-9}
    & Avg & CV & Avg & CV & Avg & CV & Avg & CV &  & \\ 
    \midrule
Michelangelo~\cite{zhao2024michelangelo} & 0.1353 & 0.7524 & 0.0229 & 0.0150 & 0.1249 & 0.0849 & 0.0898 & 0.3055 & 7.7251 & 10.9288 \\ 
CLAY~\cite{zhang2024clay}~$\dagger$ & \textbf{0.1504} & 0.8906 & 0.0063 & 0.0046 & 0.0612 & 0.0500 & 0.2218 & 0.2917 & 1.6655 & \textbf{12.2855} \\
LTM3D & 0.1478 & 0.8981 & 0.0058 & 0.0048 & 0.0579 & 0.0467 & 0.2459 & 0.3010 & 1.5878 & 11.3112 \\
LTM3D~(RGS) & 0.1487 & \textbf{0.8986} & \textbf{0.0054} & \textbf{0.0039} & \textbf{0.0555} & \textbf{0.0440} & \textbf{0.2600} & \textbf{0.3195} & \textbf{1.4239} & 11.2555 \\
\bottomrule
\end{tabular}
   \vspace{-3mm}
  \caption{Image-conditioned generation comparison on ShapeNet. `Avg' represents the average metric across 15 categories, while `CV' denotes the metric calculated across generation results using 6 different views as conditions. $\dagger$ indicates that the metrics are reproduced by us using same protocols. RGS means our Reconstruction Guided Sampling strategy.}
  \label{tab:image_cond_gen}
     \vspace{-2mm}
\end{table*}

\begin{table*}
  \centering
  \begin{tabular}{l|c|c|c|c|c|c}
    \toprule
    Methods & ULIP\,($\uparrow$) & CD\,($\downarrow$) & EMD\,($\downarrow$) & F-Score\,($\uparrow$) & P-FID\,($\downarrow$) & P-IS\,($\uparrow$) \\
    \midrule
Michelangelo~\cite{zhao2024michelangelo} & 0.1353 & 0.0370 & 0.1555 & 0.0629 & 4.1197 & \textbf{13.010} \\ 
CLAY~\cite{zhang2024clay}~$\dagger$ & 0.1343 & 0.0365 & 0.1541 & 0.0969 & 7.1178 & 12.6475 \\
LTM3D & 0.1388 & 0.0280 & 0.1333 & 0.1108(8) & \textbf{1.8556} & 12.5195 \\
LTM3D~(RGS) & \textbf{0.1421} & \textbf{0.0261} & \textbf{0.1295} & \textbf{0.1109(4)} & 2.1448 & 12.5444 \\
\bottomrule
\end{tabular}
   \vspace{-3mm}
  \caption{Text-conditioned generation comparison on ShapeNet. The results are averaged across 55 categories. $\dagger$ indicates that the metrics are reproduced by us using same protocols. RGS means our Reconstruction Guided Sampling strategy.}
  \label{tab:text_cond_gen}
    \vspace{-4mm}
\end{table*}

We evaluate the conditional generation performance of LTM3D and other state-of-the-art methods on Signed Distance Field (SDF) data, given the growing popularity of SDF-based approaches~\cite{zhang20223dilg, li2023diffusion, zhang20233dshape2vecset, zhang2024clay} in 3D shape generation. For other 3D representations, additional visualizations are provided in Sec.~\ref{sec:other_3d}, and comparisons are included in the supplementary materials. Details of the experiment implementation are provided in the supplementary materials.

\begin{figure*}[!t]
  \centering
  \includegraphics[width=1.0\linewidth]{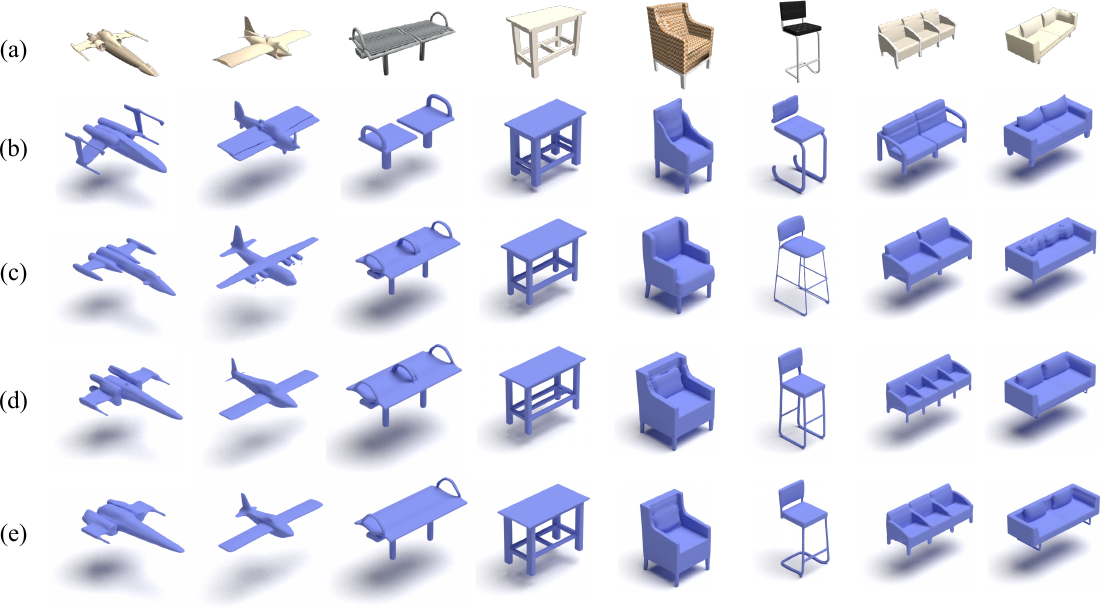}
     \vspace{-8mm}
  \caption{Qualitative image-conditioned generation results on ShapeNet. Row (a) displays the input images, row (b) shows the generation results from Michelangelo~\cite{zhao2024michelangelo}, row (c) presents results from CLAY~\cite{zhang2024clay}, row (d) demonstrates LTM3D’s generation results, and row (e) illustrates LTM3D’s results using the Reconstruction Guided Sampling strategy.}
  \label{fig:single_view}
    \vspace{-4mm}
\end{figure*}

\begin{figure*}[!t]
  \centering
  \includegraphics[width=0.88\linewidth]{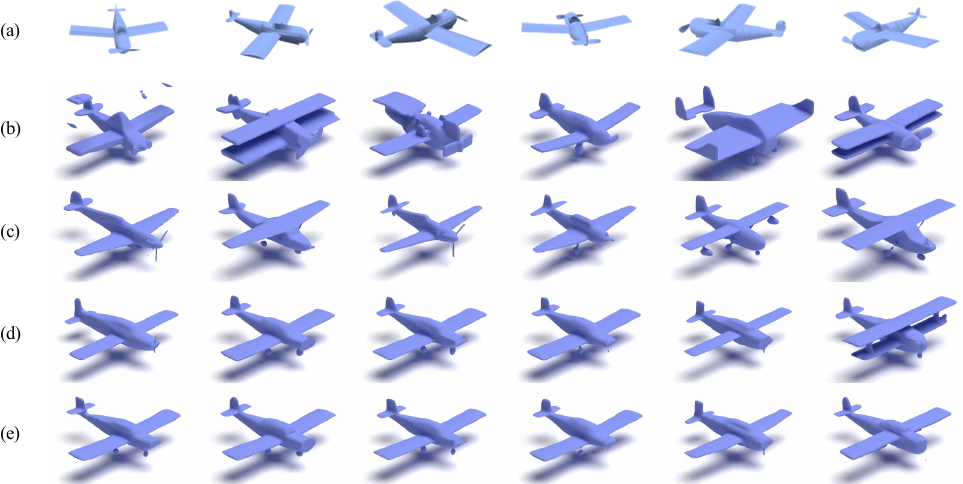}
     \vspace{-4mm}
  \caption{Image-conditioned generation results for different input views on ShapeNet. Row (a) displays the input images, row (b) shows the generation results from Michelangelo~\cite{zhao2024michelangelo}, row (c) presents results from CLAY~\cite{zhang2024clay}, row (d) demonstrates LTM3D’s generation results, and row (e) illustrates LTM3D’s results using the Reconstruction Guided Sampling strategy.}
  \label{fig:multi_view}
    \vspace{-2mm}
\end{figure*}

\begin{figure*}[!t]
  \centering
  \includegraphics[width=0.88\linewidth]{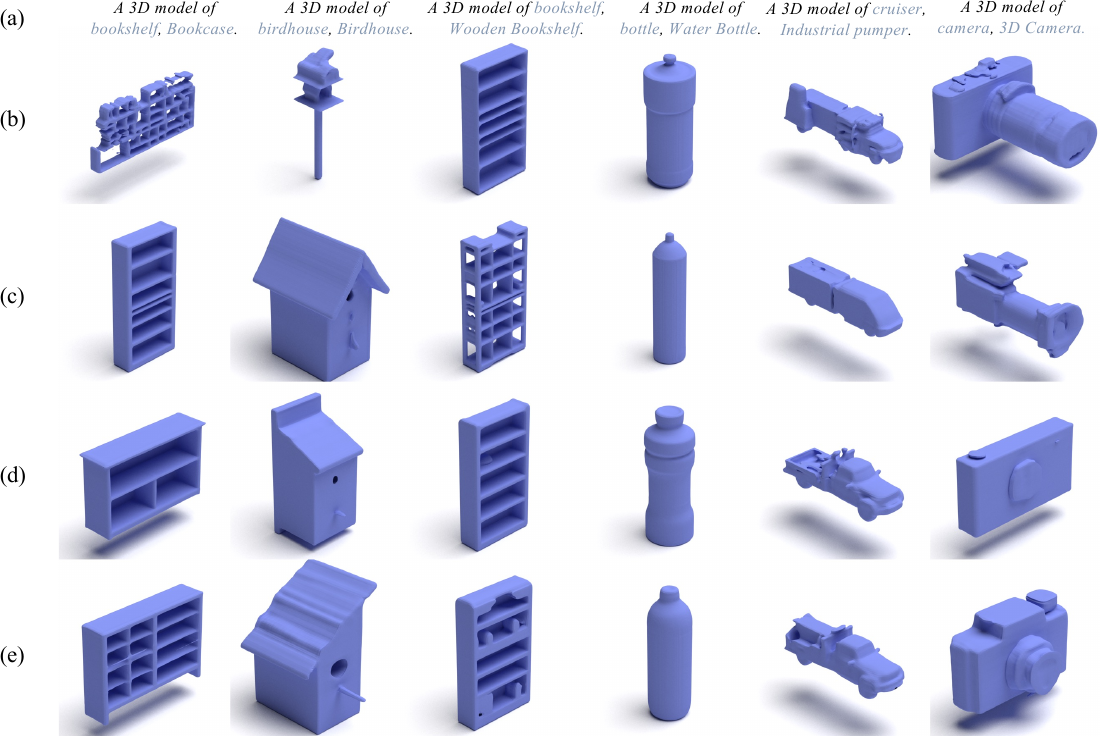}
     \vspace{-3.5mm}
  \caption{Qualitative text-conditioned generation results on ShapeNet. Row (a) displays the input text descriptions, row (b) shows the generation results from Michelangelo~\cite{zhao2024michelangelo}, row (c) presents results from CLAY~\cite{zhang2024clay}, row (d) demonstrates LTM3D’s generation results, and row (e) illustrates LTM3D’s results using the Reconstruction Guided Sampling strategy.}
  \label{fig:text view}
     \vspace{-5mm}
\end{figure*}

\begin{figure}[!t]
  \centering
  \includegraphics[width=1.0\linewidth]{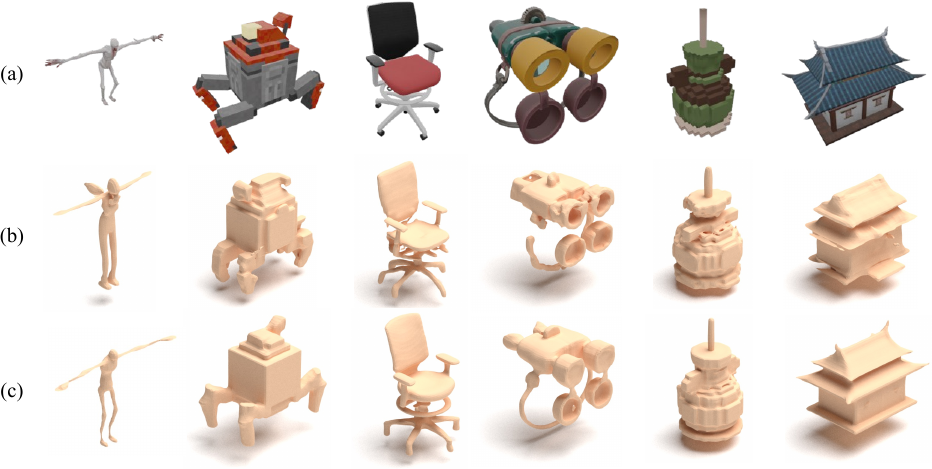}
     \vspace{-4mm}
  \caption{Image-conditioned generation results on Objaverse. Row (a) displays the input images, row (b) shows the generation results from CLAY~\cite{zhang2024clay}, row (c) demonstrates LTM3D’s generation results.}
  \label{fig:objaverse_vis}
     \vspace{-3mm}
\end{figure}

\subsection{Dataset and Metrics}
\noindent
\textbf{Dataset.} We evaluate our conditional shape generation framework using the ShapeNet dataset~\cite{chang2015shapenet} and the Objaverse dataset~\cite{deitke2023objaverse}.
For ShapeNet-based generation, we use the rendering results from DISN’s ShapeNet Rendering dataset~\cite{xu2019disn} to obtain image prompts. Additionally, we construct text prompts based on ShapeNet’s annotated labels and tags, following Michelangelo~\cite{zhao2024michelangelo}. The text prompts follow the format: “A 3D model of ${label}$, ${tag}$.”
For image-conditioned generation on Objaverse, we render six evenly distributed views for approximately 160k samples to serve as image prompts.

\noindent
\textbf{Metrics.} To assess the alignment between generated shapes and conditioning images or text descriptions, we use a pre-trained ULIP~\cite{xue2023ulip} model to compute feature similarity between 3D shapes and their corresponding image or text features, denoted as ULIP score. Additionally, we measure the generation accuracy between generated shapes and ground truth shapes using Chamfer Distance (CD), Earth Mover’s Distance (EMD), and F-score. For assessing the quality and diversity of generated shapes, we employ the P-FID and P-IS metrics as introduced in Point-E~\cite{nichol2022point}. Note that the P-FID score is calculated between the generated shapes and ground truth shapes on the test dataset. For image-conditioned generation, we further assess cross-view consistency by computing cross-view metrics using ULIP, CD, EMD, and F-score measures.

\subsection{Experimental Comparison}
We evaluate the effectiveness of LTM3D on both image- and text-conditioned SDF generation tasks. For diffusion-based models, we select two recent state-of-the-art SDF generation methods: Michelangelo~\cite{zhao2024michelangelo} and CLAY~\cite{zhang2024clay}. For AR models, we use AutoSDF~\cite{mittal2022autosdf} as a baseline for comparison.
Since CLAY’s code and training data are not available, we reproduce its results by implementing a 540M-parameter diffusion model for CLAY, while our LTM3D has 500M parameters.
The model size is determined by adjusting the number and dimension of the attention head, following CLAY’s strategy.
For a fair comparison, we use the same VAE model for SDF encoding and maintain the same training GPU hours.

\noindent
\textbf{Image conditioned generation on ShapeNet.}
Results in Tab.~\ref{tab:image_cond_gen} show that LTM3D outperforms diffusion-based methods across multiple metrics, including CD, EMD, F-Score, and P-FID, achieving state-of-the-art results. These metrics validate LTM3D's ability to produce shapes closely aligned with ground truth data.
We also achieve comparable performance to CLAY on ULIP scores, demonstrating LTM3D’s ability to generate highly faithful shapes.
The visualization results in Fig.\ref{fig:single_view} further highlight these improvements. Unlike Michelangelo\cite{zhao2024michelangelo} and CLAY~\cite{zhang2024clay}, which struggle to accurately capture structural details from input images, LTM3D precisely reconstructs the intended shapes. Notable examples include the distinct \textit{plane structure} in the second column, the \textit{table shape} in the fourth column, and the \textit{chair leg topology} in the sixth column. Additionally, our \textit{Reconstruction Guided Sampling} strategy further refines performance, leading to more faithful shape generation.
Regarding the P-IS score, our method prioritizes accuracy in generation, which may result in reduced diversity when conditioned on the same input. However, overall generation diversity can still be maintained by varying input conditions.
Furthermore, results in Tab.~\ref{tab:comp_autosdf} demonstrate that LTM3D also surpasses the AR-based framework in image-conditioned generation.

\noindent
\textbf{Image conditioned generation on Objaverse.}
We also compare the image-conditioned generation results of LTM3D and CLAY on the Objaverse dataset.
As shown in Tab.\ref{tab:comp_objaverse}, LTM3D demonstrates superior performance on this larger dataset. Additionally, the visualization in Fig.\ref{fig:objaverse_vis} highlights LTM3D’s ability to generate more faithful topologies compared to CLAY, further showcasing its capability to produce complex shapes. 
More visualization results are provided in supplementary materials.

\noindent
\textbf{Cross-view generation consistency on ShapeNet.}
To evaluate the generation consistency across different input views of a given shape, we calculate cross-view metrics using the ULIP, CD, EMD, and F-Score measures. The results in Tab.\ref{tab:image_cond_gen} show that LTM3D achieves higher consistency in generated shapes when conditioned on images from varying perspectives.
The visualization results in Fig.~\ref{fig:multi_view} illustrate the cross-view consistency of each method. Michelangelo~\cite{zhao2024michelangelo} produces varied structures for each input view, and CLAY’s~\cite{zhang2024clay} outputs (last two columns in row (c)) also show inconsistency across views. In contrast, the results in rows (d) and (e) demonstrate that our \textit{Reconstruction Guided Sampling} strategy effectively preserves stable shape structures for conditions across different viewpoints.

\begin{table}
  \centering
\resizebox{0.46\textwidth}{!}{  \begin{tabular}{c|c|c|c|c}
    \toprule
    Exp & ULIP\,($\uparrow$) & CD\,($\downarrow$) & EMD\,($\downarrow$) & F-Score\,($\uparrow$) \\
    \midrule
AutoSDF~\cite{mittal2022autosdf} & 0.1247 & 0.0189 & 0.1332 & 0.0627 \\
LTM3D & \textbf{0.1688} & \textbf{0.0048} & \textbf{0.0622} & \textbf{0.1589} \\
\bottomrule
\end{tabular}
}
  \vspace{-3mm}
  \caption{Image-conditioned generation on ShapeNet chair set.}
  \label{tab:comp_autosdf}
    \vspace{-3.5mm}
\end{table}

\begin{table}
  \centering
\resizebox{0.46\textwidth}{!}{  \begin{tabular}{c|c|c|c|c}
    \toprule
    Exp & ULIP\,($\uparrow$) & CD\,($\downarrow$) & EMD\,($\downarrow$) & F-Score\,($\uparrow$) \\
    \midrule
CLAY~\cite{zhang2024clay}$\dagger$ & 0.1619 & 0.0133 & \textbf{0.0963} & 0.1020 \\
LTM3D & \textbf{0.1694} & \textbf{0.0130} & 0.0985 & \textbf{0.1478} \\
\bottomrule
\end{tabular}
}
  \vspace{-3mm}
  \caption{Image-conditioned generation on Objaverse. $\dagger$ indicates that the metrics are reproduced by us using same protocols.}
  \label{tab:comp_objaverse}
    \vspace{-3.5mm}
\end{table}

\noindent
\textbf{Text conditioned generation on ShapeNet.}
The results for text-conditioned generation are shown in Tab.~\ref{tab:text_cond_gen}. LTM3D achieves the highest performance across ULIP, CD, EMD, F-Score, and P-FID metrics, highlighting the faithfulness and precision of our method in generating shapes that align well with the text prompts.
The visualization results in Fig.~\ref{fig:text view} further illustrate these strengths. Rows (d) and (e) show that LTM3D generates more detailed structures than Michelangelo~\cite{zhao2024michelangelo} and CLAY~\cite{zhang2024clay}, such as a \textit{bookshelf} with regular shelving sections, an \textit{Industrial pumper} with fine details and a \textit{camera} with precise contours.
However, due to the limited descriptiveness of the text prompts, our latent token reconstruction module shows slightly reduced reconstruction quality, which affects shape quality and results in a lower P-FID, despite the improved geometric accuracy. 
The P-IS score indicates that LTM3D’s diversity is limited when conditioned on same text prompts, while using more diverse text prompts can alleviate this limitation.

\begin{table}
  \centering
\resizebox{0.46\textwidth}{!}{  \begin{tabular}{c|c|c|c|c}
    \toprule
    Exp & ULIP\,($\uparrow$) & CD\,($\downarrow$) & EMD\,($\downarrow$) & F-Score\,($\uparrow$) \\
    \midrule
wo PL & 0.1452 & 0.0061 & 0.0611 & 0.2387 \\
w PL & \textbf{0.1478} & \textbf{0.0058} & \textbf{0.0579} & \textbf{0.2459} \\
\bottomrule
\end{tabular}
}
  \vspace{-3mm}
  \caption{Ablation study of Prefix Learning module.}
  \label{tab:ablation_prefix_learning}
    \vspace{-3.5mm}
\end{table}

\begin{table}
  \centering
\resizebox{0.46\textwidth}{!}{  \begin{tabular}{c|c|c|c|c}
    \toprule
    Step & ULIP\,($\uparrow$) & CD\,($\downarrow$) & EMD\,($\downarrow$) & F-Score\,($\uparrow$) \\
    \midrule
$\le$ 10 & 0.1485 & 0.0056 & 0.0566 & 0.2569   \\ 
$\le$ 20 & 0.1487 & 0.0055 & 0.0559 & 0.2592   \\ 
$\le$ 30 & 0.1487 & 0.0054 & 0.0555 & 0.2600   \\ 
$\le$ 40 & 0.1484 & 0.0053 & 0.0553 & 0.2603   \\
\bottomrule
\end{tabular}
}
  \vspace{-3mm}
  \caption{Ablation study of different fusion steps.}
  \label{tab:fusion_step_ablation}
    \vspace{-5mm}
\end{table}

\begin{table}
  \centering
  \begin{tabular}{c|c|c|c|c}
    \toprule
    Ratio & ULIP\,($\uparrow$) & CD\,($\downarrow$) & EMD\,($\downarrow$) & F-Score\,($\uparrow$) \\
    \midrule
0.1 & 0.1487 & 0.0054 & 0.0555 & 0.2600 \\
0.2 & 0.1486 & 0.0054 & 0.0554 & 0.2607  \\
0.3 & 0.1485 & 0.0054 & 0.0553 & 0.2612  \\
0.4 & 0.1485 & 0.0054 & 0.0553 & 0.2613  \\
\bottomrule
\end{tabular}
  \vspace{-3mm}
  \caption{Ablation study of different fusion ratios.}
    \vspace{-6mm}
  \label{tab:fusion_ratio_ablation}
\end{table}

\subsection{Ablation Study} \label{sec:ablation}
We have conducted ablation studies focusing on two aspects including our \textit{Prefix Leaning} module and our \textit{Reconstruction Guided Sampling} strategy.

\noindent
\textbf{Prefix Learning.} 
We conduct image-conditioned generation experiments using two approaches: (1) directly concatenating DinoV2~\cite{oquab2023dinov2} image features with the input token sequence, and (2) using our \textit{Prefix Learning} module to transfer DinoV2 image features into prefix tokens, which are then concatenated with the shape tokens. A linear layer is employed to align the dimensions of the condition tokens and shape tokens.
As shown in Tab.~\ref{tab:ablation_prefix_learning}, the results for ULIP, CD, EMD, and F-Score metrics indicate that using our \textit{Prefix Learning} module yields more faithful and precise generation results. This demonstrates that transferring the condition tokens into the same latent space as the input tokens allows for better alignment of the generated shapes with the input conditions.

\noindent
\textbf{Reconstruction Guided Sampling.}
During the auto-regressive sampling stage, we evaluate our \textit{Reconstruction Guided Sampling} strategy with varying fusion parameters.
First, we test different fusion steps, with the results presented in Tab.~\ref{tab:fusion_step_ablation}. As the fusion step increases from 10 to 40, the CD, EMD, and F-Score metrics improve, indicating a closer geometric match to the ground truth shape. However, when the fusion step reaches 40, the ULIP metric declines, suggesting a drop in shape quality. To balance generation quality and accuracy, we selected a fusion step of 30 as the optimal setting.
Similarly, based on the results in Tab.~\ref{tab:fusion_ratio_ablation}, we determine 0.1 to be the ideal fusion ratio, providing a balance that preserves both generation quality and accuracy.

\subsection{3D Gaussians, Point Cloud and Mesh}
\label{sec:other_3d}
Our LTM3D framework supports various 3D shape representations by integrating corresponding shape encoders.
We conduct image-conditioned generation experiments on 3DGS, point cloud, and mesh data. More comparisons, additional visualizations, and detailed implementation are provided in the supplementary material. As shown in Fig.~\ref{fig:gs_pcd_mesh}, these results demonstrate LTM3D’s capability for conditional shape generation across diverse data representations, highlighting its versatility and effectiveness in adapting to different 3D data formats.

\begin{figure}[h]
    \centering
    \includegraphics[width=1\linewidth]{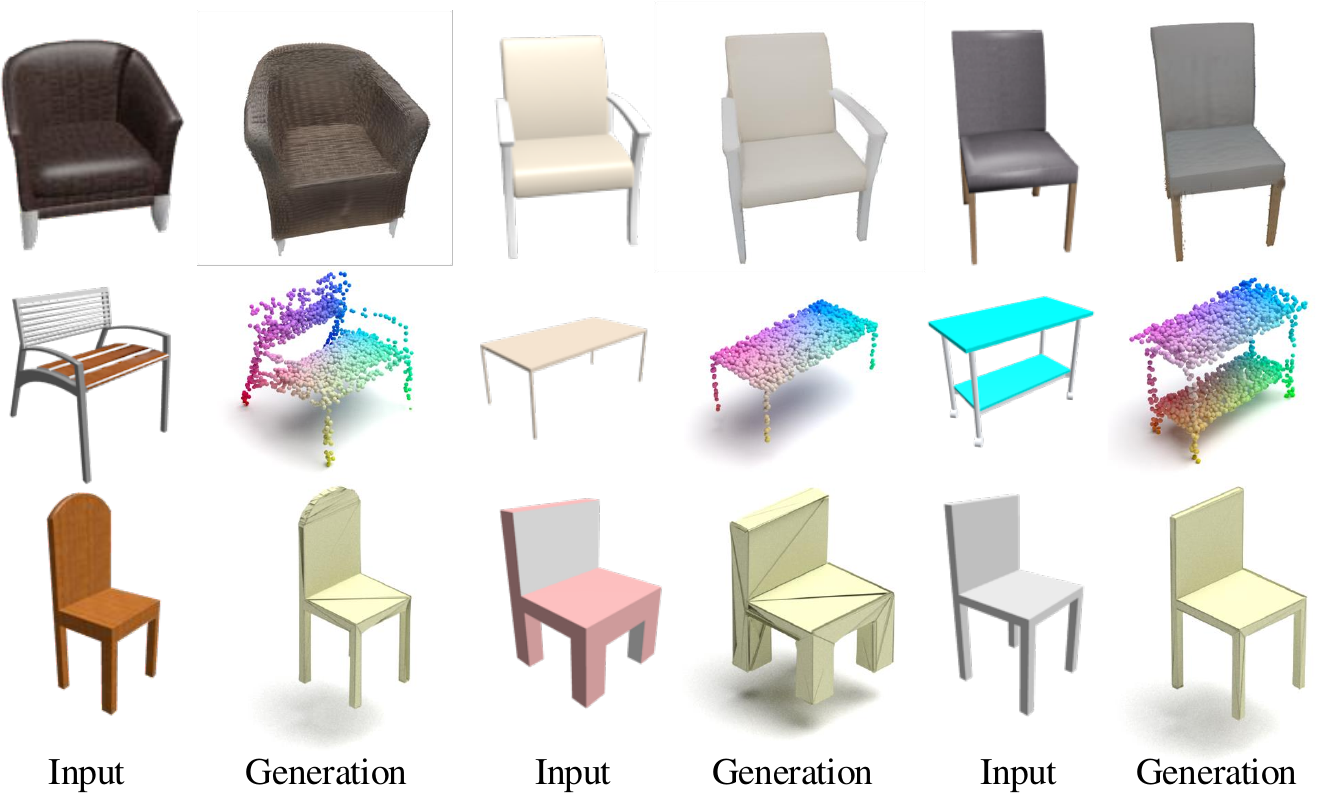}
      \vspace{-4mm}
    \caption{Qualitative image-conditioned generation results on ShapeNet using various 3D representations including 3D Gaussian Splatting, point cloud and mesh.}
    \label{fig:gs_pcd_mesh}
     \vspace{-8mm}
\end{figure}


\section{Limitations and Future Work} \label{section:limitation}
While our LTM3D provides a latent token space modeling framework for shape generation that supports different 3D data representations, it still requires training a separate model for each representation.
This limitation arises because the latent spaces for different 3D data types are not aligned during encoding. Attempts to jointly train with mixed representations have yielded suboptimal results, as the encoding of different data formats does not integrate seamlessly. A promising direction for future research is to develop a unified latent space capable of accommodating multiple data formats, enabling a single model to generate diverse 3D representations effectively. 

\section{Conclusion} \label{section:conclusion}

In this work, we present LTM3D, a latent token space modeling framework for conditional 3D shape generation designed to support multiple 3D data representations and multi-modal prompts. 
To enhance prompt fidelity, we introduce a \textit{Prefix Learning} module that aligns image or text tokens with the shape latent space, reducing the domain gap and improving prompt adherence. We also propose a \textit{Latent Token Reconstruction} module that maps conditional prompts to shape latent tokens, which serve as priors in a \textit{Reconstruction Guided Sampling} strategy. This helps to stabilize the sampling process and ensures more accurate shapes that align closely with conditioning prompts.
In summary, LTM3D marks a significant advancement toward a general pre-training model for 3D shape generation, capable of seamlessly adapting to various 3D data formats while maintaining high fidelity and structural coherence in generated shapes.

{
    \small
    \bibliographystyle{ieeenat_fullname}
    \bibliography{main}
}

\clearpage
\setcounter{page}{1}
\maketitlesupplementary
\renewcommand{\thesection}{\Alph{section}} 
\setcounter{section}{0} 

\begin{figure*}[!t]
  \centering
  \includegraphics[width=1.0\linewidth]{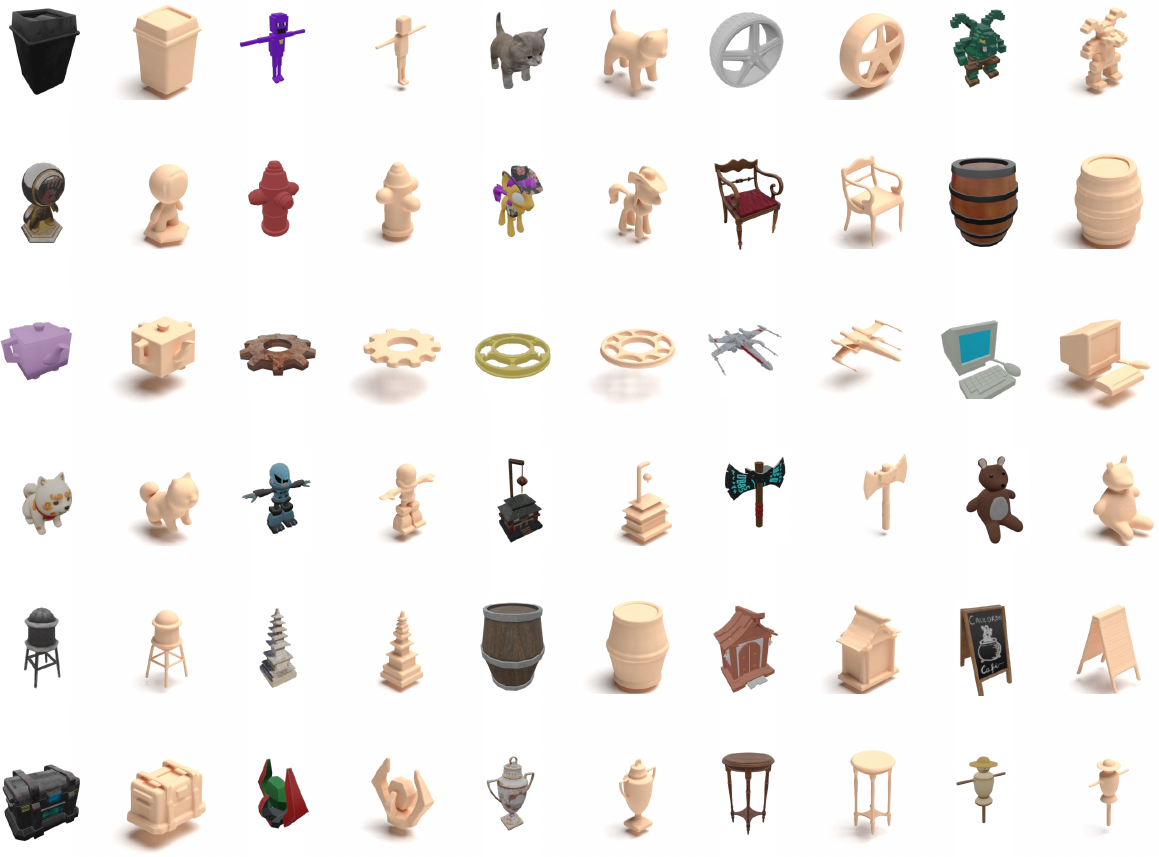}
  \vspace{-6mm}
  \caption{Image-conditioned generation results for SDF data on Objaverse. Each pair consists of an input image (left) and the corresponding generation result (right).}
  \vspace{-4mm}
  \label{fig:supp_vis_objaverse}
\end{figure*}

\begin{figure*}[!t]
  \centering
  \includegraphics[width=1.0\linewidth]{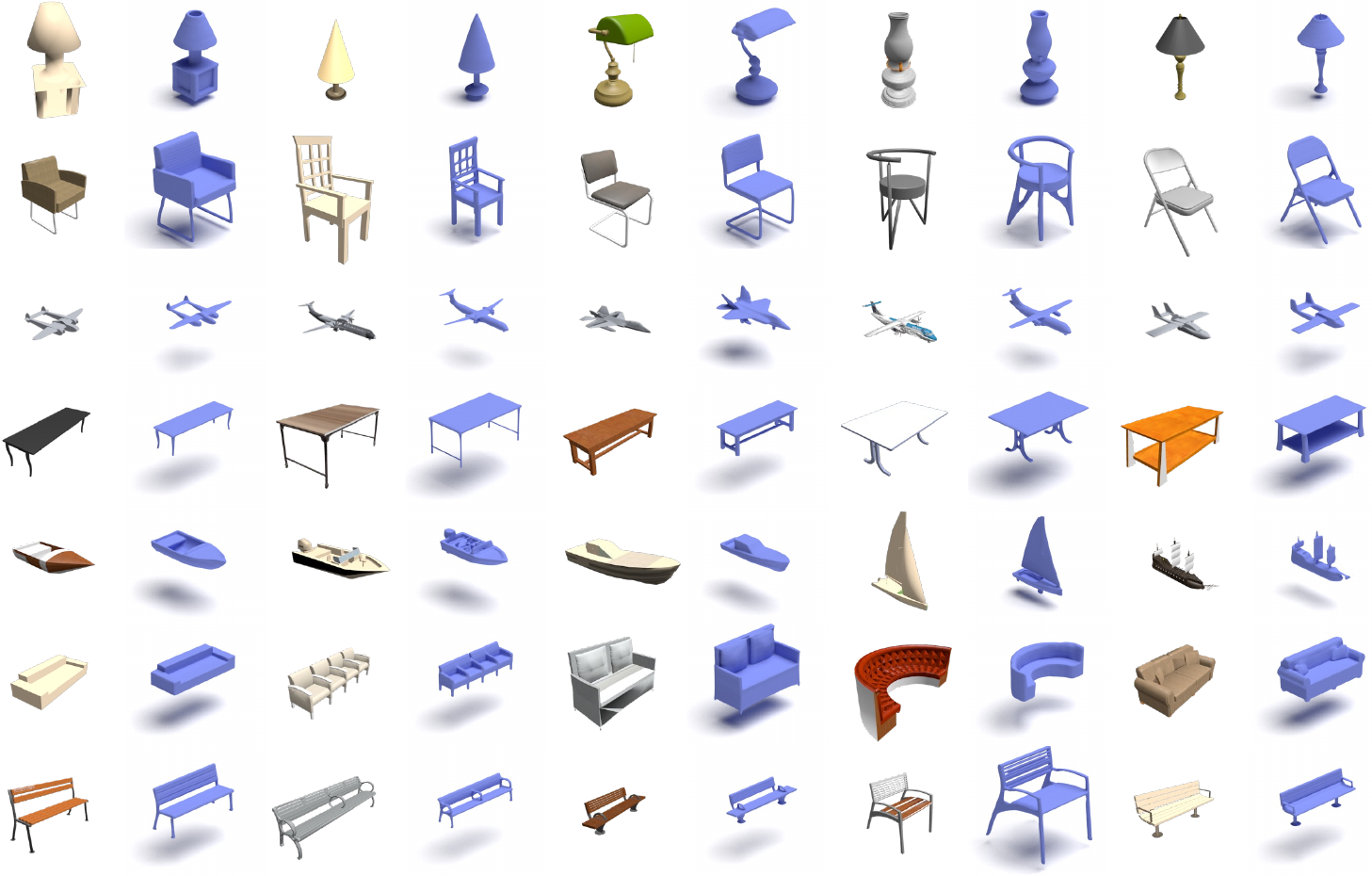}
  \vspace{-6mm}
  \caption{Image-conditioned generation results for SDF data on ShapeNet. Each pair consists of an input image (left) and the corresponding generation result (right).}
  \vspace{-4mm}
  \label{fig:supp_vis_sdf}
\end{figure*}

\section{Ethics Statement}

This work is conducted as part of a research project. While we plan to share the code and findings to promote transparency and reproducibility in research, we currently have no plans to incorporate this work into a commercial product. In all aspects of this research, we are committed to adhering to Microsoft AI principles, including fairness, transparency, and accountability.

\section{Implementation Details}

\subsection{Shape encoding}\label{sec:supp_shape_enc}
For different 3D representations, we employ distinct encoding strategies. To demonstrate the effectiveness of our pipeline, we keep the encoding strategy simple and straightforward, adopting methods used in existing works~\cite{zhang20233dshape2vecset, siddiqui2024meshgpt, nichol2022point}.
The details are described as below.

\noindent
\textbf{SDF encoding.} To encode signed distance field (SDF) data, we utilize the variational auto-encoder (VAE) framework from 3DShape2VecSet~\cite{zhang20233dshape2vecset}. To initialize the shape latent codes, we employ a learnable query set consisting of 512 latent vectors, each with a dimensionality of 512.
The encoder and decoder modules are used with their default configurations, and the dimensionality of the variational latent codes is set to 8. Training is conducted over 200 epochs with a learning rate of $1 \times 10^{-4}$ and a batch size of 64. The training objective is cross-entropy loss following 3DShape2VecSet.

\noindent
\textbf{3DGS encoding.} 
We follow GaussianCube~\cite{zhang2024gaussiancube} to represent each 3D Gaussian data as a volume with dimensions $\mathcal{R}^{32^3\times8}$. Here, $32$ denotes the resolution of the volume, and $8$ corresponds to the dimensionality of the attributes. To compress the Gaussian volume into a learnable query set, we adopt the same VAE model as used in 3DShape2VecSet. The VAE architecture and training setup are consistent with those used for SDF encoding. The training objective consists of a Mean Squared Error (MSE) loss for shape reconstruction and an image rendering loss for high-quality rendering, as proposed in GaussianCube.

\noindent
\textbf{Point cloud encoding.}
For point cloud data, we represent each shape using $xyz$ coordinates as shape tokens, sampling 1,024 points per shape during training. The coordinates are normalized to the range $[-1, -1, -1]$ to $[1, 1, 1]$. After generation, we apply Point-E’s up-sampling model~\cite{nichol2022point} to produce the final point cloud with 4,096 points.

\noindent
\textbf{Mesh encoding.}
For mesh data, we utilize face sequences processed by MeshGPT~\cite{siddiqui2024meshgpt}, where each face is defined by three vertices. The vertex coordinates of each face are concatenated to form mesh tokens, resulting in a sequence with dimensions $\mathcal{R}^{n \times 9}$, where $n$ represents the number of faces. To enable batch processing of meshes with varying numbers of faces, we repeat each face sequence multiple times and truncate the sequence to the first $N$ tokens during training. In our experiments, $N$ is set to 800.


\subsection{Prefix Learning} For image prompts, we use DinoV2~\cite{oquab2023dinov2} to generate condition tokens with the shape $\mathcal{R}^{(256+1) \times 1536}$, where 256 denotes the number of patch features and 1 represents the global feature. For text prompts, we employ CLIP’s~\cite{radford2021learning} text encoder to produce a condition token of shape $\mathcal{R}^{768}$. To convert condition tokens into prefix tokens, we initialize a set of learnable queries $\bm{Q}$ with the shape $\mathcal{R}^{64 \times d}$ and apply a cross-attention layer followed by a feed-forward layer to generate the prefix tokens, setting the hidden feature dimension as $1,024$.

\subsection{Conditional Distribution Modeling} Conditional Distribution Modeling comprises a \textit{Masked Auto-encoder} and a \textit{DenoiseNet}.
For the MAE~\cite{he2022masked} model, we follow MAR~\cite{li2024autoregressive}, employing two 16-layer transformer blocks as the encoder and decoder, respectively, with a hidden feature dimension of 1,024. During training, the mask ratio is randomly sampled within the range [0.7, 1.0].
Similarly, we construct the DenoiseNet following MAR, using 8 MLP layers with a hidden feature dimension of 1,280.
Training is guided by the diffusion loss.

\subsection{Latent Token Reconstruction}
To reconstruct latent tokens from image or text conditions, we design a latent token reconstruction module consisting of a cross-attention layer followed by 24 self-attention layers. The number and dimension of learnable tokens are set to 512 and 512, respectively. We use Mean Squared Error loss as the training objective for this module.

\subsection{Reconstruction Guided Sampling}
During the auto-regressive sampling process, we set a fusion weight $\alpha_i$ to linearly blend the reconstructed tokens and sampled tokens to generate condition $\bm{z}_i$.
Since the sampling randomness decays as the sequence length increases, we set $\alpha_i=0.1$ for $i\le 30$.

\section{Training details}

\subsection{LTM3D training} \label{sec:supp_train}
For the Prefix Learning module, we define prefix tokens with dimensions $\mathcal{R}^{64 \times 1024}$ to serve as conditioning inputs. For image-conditioned generation, we utilize DinoV2 image features with dimensions $\mathcal{R}^{257 \times 1536}$, while for text-conditioned generation, we use CLIP text embeddings with dimensions $\mathcal{R}^{1 \times 768}$. The module consists of a cross-attention layer and a feed-forward layer with a hidden dimension of 1,024. An additional linear layer is employed to project the dimension of condition tokens into 1,024.
For the masked auto-encoder (MAE), we adopt the same architecture as MAR~\cite{li2024autoregressive}, comprising a 16-layer encoder and a 16-layer decoder with a hidden dimension of 1,024. For DenoiseNet, we follow the MAR architecture, utilizing 8 MLP layers with a hidden dimension of 1,280.
The entire framework is trained for 800 epochs with a batch size of 256 and a learning rate of $1 \times 10^{-4}$. The training objective is the diffusion loss.
We use AdamW as the optimizer.
For 3DGS data, we further incorporate an image rendering loss during training, as proposed in GaussianCube~\cite{zhang2024gaussiancube}, to ensure the rendering quality of the sampled 3D Gaussians.

\subsection{Latent Token Reconstruction training}
To reconstruct latent tokens from input condition tokens, we define a learnable query set with dimensions $\mathcal{R}^{n \times d}$, where $n$ represents the number of shape tokens and $d=512$ is the vector dimension. DinoV2 image features and CLIP text embeddings are used as the condition tokens.
The module comprises a cross-attention layer followed by 24 self-attention layers, each with a hidden dimension of 512. It is trained for 200 epochs using AdamW as the optimizer, with a learning rate of $1 \times 10^{-4}$ and a batch size of 64. The training objective is Mean Squared Error (MSE) loss.

\section{Inference details}
We follow the same inference process as MAR~\cite{li2024autoregressive}, which involves iteratively sampling tokens in a random order. However, during the first 30 sampling steps, we introduce a modification: the sampled tokens are blended with our reconstructed tokens using a linear weight of $\alpha=0.9$. The formulation is presented in Eq.~4. Importantly, this blending is only used for generating conditions within the MAE module and does not alter the final sampled tokens. Remaining inference settings are identical to those of MAR.

\begin{figure*}[h]
  \centering
  \vspace{-4mm}
  \includegraphics[width=1.0\linewidth]{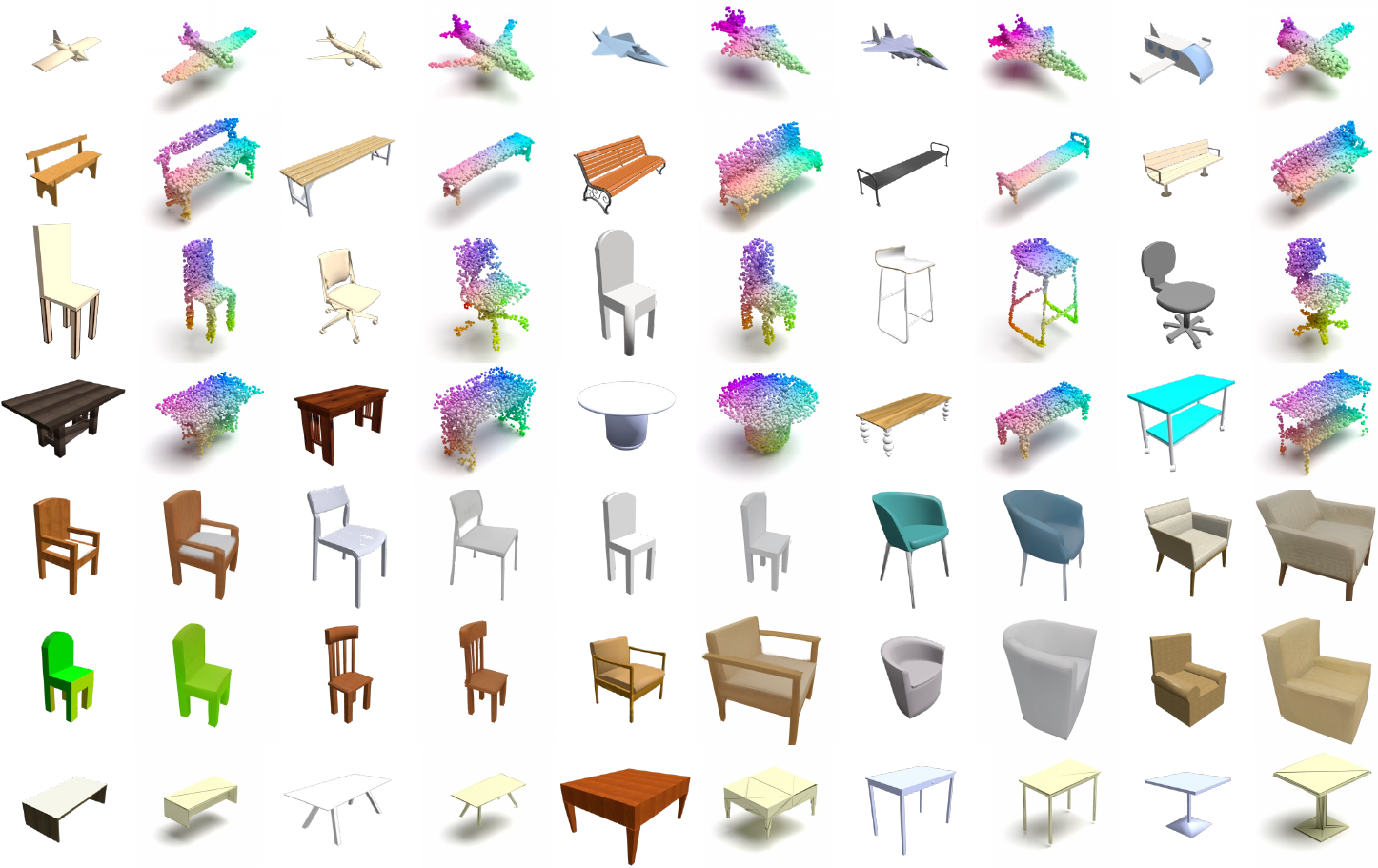}
  \vspace{-6mm}
  \caption{Image-conditioned generation results for point cloud, 3DGS, and mesh data on ShapeNet. Each pair includes an input image (left) and the corresponding generation result (right). The first four rows display results for point cloud data, the fifth and sixth rows show results for 3D Gaussian data, and the final row presents results for mesh data.}
  \label{fig:supp_vis_other}
  \vspace{-4mm}
\end{figure*}

\section{More visualization and comparison results}

We present additional visualization results for image-conditioned generation in Fig.\ref{fig:supp_vis_objaverse}, Fig.\ref{fig:supp_vis_sdf} and Fig.\ref{fig:supp_vis_other}. The results in Fig.\ref{fig:supp_vis_objaverse} and Fig.\ref{fig:supp_vis_sdf} emphasize the faithful generation capabilities of our LTM3D framework, showcasing the effectiveness of the Prefix Learning module and the Reconstruction Guided Sampling strategy. Meanwhile, the results in Fig.\ref{fig:supp_vis_other} demonstrate that LTM3D can seamlessly handle diverse 3D representations, further supporting its role as a general framework for conditional 3D shape generation.
While LTM3D achieves high faithfulness and quality in generating diverse 3D representations, some limitations remain. For instance, the generated colors for 3DGS data exhibit slight deviations from the input image, and the generated meshes display minor artifacts. These imperfections result from the use of straightforward encoding methods that lack specialized design, limiting the expressiveness of the shape tokens.
This further implies the importance of developing more effective shape encoding strategies.

We also evaluate our LTM3D for 3DGS and point cloud generation separately on the ShapeNet Chair subset and the ShapeNet test set. The results are presented in Tab.~\ref{tab:comp_3dgs} and Tab.~\ref{tab:comp_pcd}.
Without any specialized design or meticulous tuning, we surpass Point-E in point cloud generation and achieve results comparable to GaussianCube.

\begin{table}[h]
  \centering
    \begin{tabular}{l|c|c|c}
      \toprule
      Methods & PSNR ($\uparrow$)& SSIM ($\uparrow$) & LPIPS ($\downarrow$)\\
      \midrule
      GaussianCube & 35.13 & 0.88 & 0.13 \\
      LTM3D & 35.34 & 0.88 & 0.14 \\
      \bottomrule
    \end{tabular}
  \caption{Comparison of image-conditioned 3DGS generation on the ShapeNet chair test set.}
  \label{tab:comp_3dgs}
\end{table}

\begin{table}[h]
  \centering
    \begin{tabular}{l|c|c}
      \toprule
      Methods & Point-E & LTM3D \\
      \midrule
      CD ($\downarrow$) & 0.087 & 0.057 \\
      \bottomrule
    \end{tabular}
  \caption{Comparison of image-conditioned point cloud generation on the ShapeNet test set.}
  \label{tab:comp_pcd}
\end{table}

\section{Multi-representation joint training}
\begin{figure}[h]
    \centering
\includegraphics[width=0.9\linewidth]{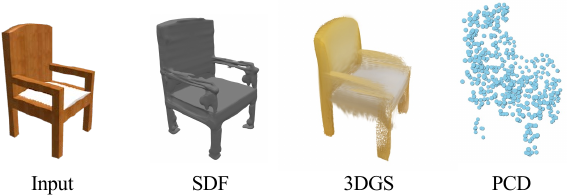}
    \caption{Visualization of joint training on ShapeNet chairs.}
    \label{fig:supp_vis_joint}
\end{figure}

We conduct experiments to jointly train an image-conditioned generation model for SDF, 3DGS, and point cloud data. Each data representation is encoded using a separate encoder as described in Sec.\ref{sec:supp_shape_enc}.
However, since these encoders are trained independently, the resulting latent spaces are not shared across representations, posing a significant challenge for training a unified generation model.
In our experiment, the number and dimension of latent tokens are uniformly set to 512 and 8. For point clouds, we sample 512 points as training data and use a simple VAE module, consisting of two linear layers, to project the $xyz$ tokens into latent tokens. To differentiate between data representations, we assign three groups of learnable position embeddings, one for each type of latent token. Furthermore, we implement three separate Prefix Learning modules while sharing backbone modules, including MAE and DenoiseNet, across all representations. The training settings are consistent with those described in Sec.~\ref{sec:supp_train}.
The results are shown in Fig.~\ref{fig:supp_vis_joint}. Since the latent tokens are encoded separately from different 3D representations, the inherent gaps between their latent spaces pose significant challenges during training, leading to low-quality generated shapes. This highlights the importance of further aligning the latent spaces of different 3D representations, which we plan to address in future work.

\end{document}